\documentclass[letterpaper, 10 pt, conference]{ieeeconf}

\IEEEoverridecommandlockouts 
\usepackage[utf8]{inputenc}
\usepackage{amsmath}
\usepackage{graphicx}
\usepackage{subcaption}
\usepackage{multirow}
\usepackage{scrextend}
\usepackage[table,xcdraw]{xcolor}
\usepackage{bm}
\usepackage{epsfig,psfrag,graphicx,color}
\usepackage{hyperref}
\usepackage{comment}

\usepackage[percent]{overpic}

\usepackage{color}
\usepackage{xcolor,colortbl}

\usepackage{amssymb}
\usepackage{afterpage}
\usepackage{pdflscape}
\usepackage{enumerate}
\usepackage{booktabs}
\let\labelindent\relax
\usepackage{enumitem}

\usepackage{amsmath}
\usepackage{algorithm}
\usepackage[noend]{algpseudocode}

\usepackage{adjustbox}
\usepackage{array}

\usepackage{cite}
\usepackage{ulem}

\newcolumntype{R}[2]{%
	>{\adjustbox{angle=#1,lap=\width-(#2)}\bgroup}%
	l%
	<{\egroup}%
}
\newcolumntype{P}[1]{>{\raggedright\arraybackslash}p{#1}}
\graphicspath{{./figs/}}

\definecolor{darkpastelgreen}{rgb}{0.01, 0.75, 0.24}
\newcommand{\IG}[1]{\textcolor{black}{#1}}

\newcommand{\corrections}[1]{\textcolor{red}{#1}}

\newif\ifshow % toggle true or false based on if want to hide section
\showfalse % hide the sections

\ifshow
  \includecomment{wrap}
\else
  \excludecomment{wrap} % anything wrapped in a {wrap} environment will be excluded
\fi

\begin{document}

\title{Standardization of Cloth Objects and its Relevance in Robotic Manipulation} 

\author{Irene Garcia-Camacho$^{1}$, Alberta Longhini$^{2}$, Michael Welle$^{2}$, Guillem Alenyà$^{1}$,  Danica Kragic$^{2}$,  Júlia Borràs$^{1}$
\thanks{$^{1}$Institut de Robòtica i Informàtica Industrial, CSIC-UPC, Barcelona, Spain
{\tt\small \{igarcia, galenya, jborras\}@iri.upc.edu}}%
\thanks{${}^2$Robotics, Perception and Learning Lab, EECS at KTH Royal Institute of Technology Stockholm, Sweden
{\tt\small \{albertal, mwelle, dani\}@kth.se}}
\thanks{\scriptsize This work has been supported by project PID2020-118649RB-I00 funded by MCIN/ AEI /10.13039/501100011033, by project PCI2020-120718-2 funded by MCIN/ AEI /10.13039/501100011033 and by the "European Union NextGenerationEU/PRTR", by CSIC iMOVE 2023 grant (code 23254) and by the European Research Council (ERC-BIRD), Swedish Research Council and Knut and Alice Wallenberg Foundation, the National Science Foundation under NSF CAREER grant number: IIS-2046491. }
}

\maketitle

\begin{abstract}

The field of robotics faces inherent challenges in manipulating deformable objects, particularly in understanding and standardising fabric properties like elasticity, stiffness, and friction. While the significance of these properties is evident in the realm of cloth manipulation, accurately categorising and comprehending them in real-world applications remains elusive. This study sets out to address two primary objectives: (1) to provide a framework suitable for robotics applications to characterise cloth objects, % quantifying its cloth properties 
and (2) to study how these properties influence robotic manipulation tasks. Our preliminary results validate the framework's ability to characterise cloth properties and compare cloth sets, and reveal the influence that different properties have on the outcome of five manipulation primitives. We believe that, in general, results on the manipulation of clothes should be reported along with a better description of the garments used in the evaluation. This paper proposes a set of these measures. 

\end{abstract}

% Keywords appear just beneath the abstract. Use only for final version. 
%\begin{IEEEkeywords}
%
%\end{IEEEkeywords}

%\IEEEpeerreviewmaketitle

\section{Introduction}\label{sec:introduction}

Manipulating deformable objects presents a significant set of unresolved challenges in robotics, encompassing modelling, perception, control and dataset standardisation~\cite{zhu2022challenges, arriola2020modeling, jimenez2020perception, garcia2020benchmarking}. Recent attention in the robotics community has been directed towards understanding fabric properties such as elasticity, bending, and friction, especially in the context of cloth manipulation~\cite{petrik2019feedback, longhini2023edonet, verleysen2020tactile}. Nonetheless, fundamental research questions remain open: How to construct comprehensive datasets for benchmarking real-world applications? What are the most influential cloth properties for manipulation? And to what extent do we need precise property identification? 
%However, the effective generalization and benchmarking of manipulation techniques across various Cloth-like Deformable Objects (CDO) with diverse mechanical and physical properties remains an underexplored area of research. Fundamental scientific questions such as how to design comprehensive datasets, what are the most influential properties for manipulation, and to what extend we need precise property identification are still open research problems. 

%We are interested in providing means to compare different datasets containing cloth-like objects and evaluate the influence of different properties when manipulating. Our aim is to advance in the standardisation and comparison between different approaches.
The focus of this work concentrates on two objectives: providing methods to characterise a cloth-like deformable object (CDO) based on its properties; and investigating the influence of such properties on manipulation tasks. %Previous work discussed variations in objects' visual appearance such as type, size and texture~\cite{muratore2022robot, matas2018sim}, while limited progress has been achieved regarding properties like elasticity, friction, and stiffness.
The properties of CDO range from physical attributes such as size, shape or color, to mechanical properties like elasticity, friction, and stiffness~\cite{grishanov2011structure}. Real-world labelled datasets usually characterise objects based on type, shape and   size~\cite{liu2016deepfashion,bertiche2020cloth3d}. Mechanical properties, instead, are often not included due to the challenges of inferring them from visual observation, being necessary physical interactions. %.  Their perception demands distinct physical interactions and specific sensor modalities, underscoring the complexities inherent in addressing 1). %requiring specific physical interactions and sensor modalities to be perceived highlighting the challenge of addressing 1). 
To avoid this limitation,  in \cite{longhini2021textile} the authors proposed a taxonomy of textiles anchored on the material and construction techniques.
%These factors influence the cloth properties leading to, for example, more elastic or less elastic behaviours. \AL{Is an example needed?} For instance, a cotton knitted fabric is usually more elastic than a cotton woven one due to the tighter weave pattern. 
Although these properties influence mechanical properties, they fail to provide quantitative measures of these. The textile industry provides standardized methods for the characterisation of elasticity, friction and stiffness. Yet, widely used tests like KES~\cite{kawabata1980KES} and FAST~\cite{giorgio1995fast} are destructive, making them unsuitable for robotic applications. 

\begin{figure}
        \centering
        \includegraphics[width=\linewidth]{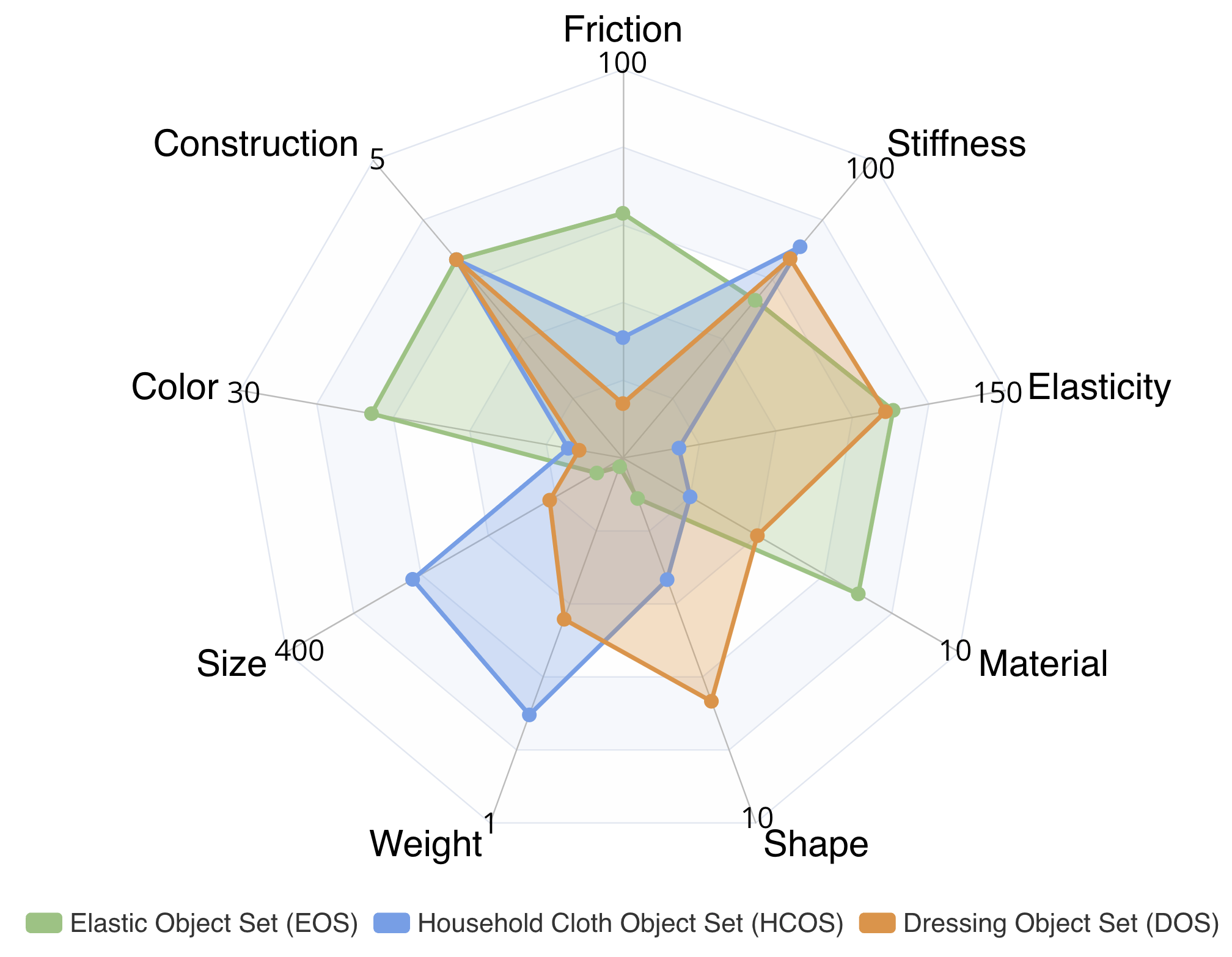}
        \caption{Representation of literature cloth sets \cite{longhini2023edonet, irene2022household, gustavsson2022landmark} based on the physical and mechanical cloth properties.}
        \label{fig:radar_chart}
        \vspace{-3pt}
\end{figure}

In this study, we address both objectives. Firstly, we equip the community with a measuring framework grounded in the textile industry's standard practices. This framework offers a set of easy-to-use measurement systems for both physical and mechanical cloth properties, allowing one to label individual cloth properties and compare different cloth sets. Secondly, leveraging the labels from our proposed framework, we conduct a novel analysis of how the properties of CDO (stiffness, elasticity and friction) influence robotic manipulations.

Our experimental results demonstrate the framework's capability of annotating cloth sets through their properties, representing them on a radar chart, and assessing the diversity of real-world object sets (Fig. \ref{fig:radar_chart}). 
In addition, preliminary findings from analysing the influence of CDO properties indicate that cloth properties are highly intertwined, but some may have more impact in certain manipulation primitives. Specifically, stiffness is a predominant property influencing all evaluated manipulation primitives. Elasticity, on the other hand, becomes critical when the cloth undergoes stress, while friction is distinctly influential during interactions between the cloth and various surfaces.

We expect our tools to improve standardized assessments of clothes, allow the creation of better cloth sets and foster benchmarking for improving cloth manipulation.

\section{Related work}\label{sec:soa}

We discuss the related work from two perspectives: cloth manipulation and textile engineering. Firstly, we identify cloth manipulation tasks, objects and related cloth properties commonly used in deformable object manipulation. Secondly, we will search for the current measurement systems used to assess these properties in textile engineering.

\subsection{Cloth manipulation} \label{sec:cloth_manip_soa}

Cloth manipulation presents many challenges due to the complexity of representing, understanding and predicting the behaviour of cloth. However, a complete understanding of the whole task is not necessarily required for accomplishing it. In \cite{irene2022knowledge} it is presented a compact and simplified definition of tasks that allows representing them as graphs described by sequences of states and transitions. The transitions of these graphs are the manipulation primitives necessary to change the state of the cloth. In \cite{borras2020graspingcentered}, they identified the most tackled tasks in literature, recognising among them unfolding, folding and flattening. These tasks comprise commonly used primitives such as drag, fold, lift, push and place \cite{avigal2022speed, ha2021flingbot, fahantidis1997robot, li2018model}. %Some works that use this way of encoding whole manipulation tasks in more simple actions are \cite{avigal2022speed, ha2021flingbot, fahantidis1997robot, li2018model}. 

In recent years, many studies in cloth manipulation and perception have used a set of objects to test their methods, indirectly showing the generalization of the method. However, these sets are often small and lack information about the objects' variability. 
In \cite{tirumala2022learning} they reported the thickness \corrections{of} the objects used but also mentioned surface texture and stiffness without providing any measurements.
System's generalization to unseen garments of different colour, shapes, and stiffness was claimed in \cite{avigal2022speed}, yet their real-world experiments involved only a limited set of objects (2 T-shirts and 1 towel). Contrarily, \cite{yuan2018} took a comprehensive approach by employing objects from various categories and materials, but classified them based on a subjective ranking of properties such as thickness, stiffness, material, stretchiness, etc. 
Several studies have attempted to assess specific cloth properties more rigorously. Li et al. \cite{li2018model} conducted experiments in a real environment to measure shear resistance and frictional forces, aiming to transfer the same cloth behaviour to simulation. Kabaya \cite{kabaya1998service} performed a test to measure fabric "softness," but the details of the procedure remain unclear. It is worth mentioning, that some works such as \cite{shibata2008handling}, applied the Kawabata's Evaluation System (KES) \cite{kawabata1980KES}, a systematic standard approach to measure in detail various cloth properties. However, this system requires specific machines that are not accessible to most researchers.
This makes it challenging to replicate results or quantify the method's generalization. Efforts have been made to establish object sets for benchmarking manipulation \cite{calli2015ycb, irene2022household, clark2023benchmark}. Yet, maintaining the exact same objects, especially with textiles, is difficult due to stock variability. To address this, it's essential to establish a standardized approach by reporting the properties that characterize textile objects.

These prior works demonstrate that there is interest in identifying and assessing textile properties,  needed to validate the generalisation of the proposed results. However, most of these works do not report quantitative measures of these parameters due to the lack of useful and effective measurement systems.

\subsection{Textile engineering} \label{sec:textile_engin_soa}%Textile industry

The textile engineering sector specializes in understanding and measuring the properties of textiles. To do so, specialized testing equipment and techniques are used to measure them accurately.%This means it has a long experience in designing measurement systems to measure different properties of fabrics.

\textbf{Stiffness:} Standard measurement systems for stiffness are the Pierce's cantilever \cite{pierce1930} and KES \cite{kawabata1980KES}. Nonetheless, these techniques are sensitive to fabric weight and edge-related influences. The Cusick drape test \cite{cusick1968}, quantifies the stiffness through its drapeability behaviour, suspending a fabric sample in a circular plate and sensing the shape of the shadow of the draped cloth projected using a light source. 

\textbf{Friction:} Regarding friction, the KES system \cite{kawabata1980KES} is also used to measure the coefficient of friction placing the fabric on a standard surface and applying a known force. Applying a known force requires specific tools, which does not reflect our goal to provide an easy-to-use system with simple tools. %This could be done \corrections{in the robotics community} using a manipulator and a F/T sensor, but our goal is to provide easy-to-use systems with simple tools.
On the contrary, with the inclined plane test \cite{mercier1930friction} the friction is measured using only a flat surface and inclining the plane.

\textbf{Elasticity:} The tensile test \cite{ASTM} is used to measure stretchability by subjecting the cloth to a controlled force. This method can also be performed manually by stretching the sample until it offers resistance. However, these approaches always use cut strips of cloth. 
We propose a set of systems that do not require damaging the textiles in any way.

\section{Textile characterization} \label{sec:properties} %\section{Measurement systems}

We propose a framework to characterise textile objects through a set of measurement systems.
The proposed systems are grounded in the methods used in textile engineering, adapted so that they can be used in robotic manipulation.
The goal is to 
provide information about how the entire object will behave in cloth manipulation tasks.
We followed the criteria: 
\begin{itemize}
    \item Non-destructive: The measurement methods do not require cutting or damaging the object in any way.
    \item General: Objects of different categories can be used, including household or clothing objects. Independently of their wear or features. %The properties tackled can be measured considering objects of different categories, including household or clothing objects. %versatile
    \item Easy to use: The proposed methods are easy to perform with simple tools. % We have seek to reuse metrics/instrucitoins (e.g. size and elasticity lines), and avoid excepetions in the methods to keep it simple (i.e. avoid different instructions depending on the object (e.g. same lines for T-shirt and top)) %User-friendly
    %\item Simple tools: When required, the hardware or tools used are easy to acquire and inexpensive. %Basic toold %tools commonly found
    \item Repeatable: The systems have clear procedures that can be consistently repeated to obtain the same results. 
\end{itemize}

Check the related website for more details of the measurement systems \footnote{\fontsize{6}{8} \label{fn:website}\url{http://www.iri.upc.edu/groups/perception/\#ClothStandardization}}.

\subsection{Physical properties}

The physical properties are inherent characteristics of the object and can be observed and measured without subjecting the material to external forces or manipulations. These properties describe how a material appears under static or non-changing conditions. %It includes its size, shape and colour, as well as the fabric material and construction technique when it comes to textiles.

\begin{figure}
\vspace{4pt}
        \centering
        \includegraphics[width=0.5\linewidth]{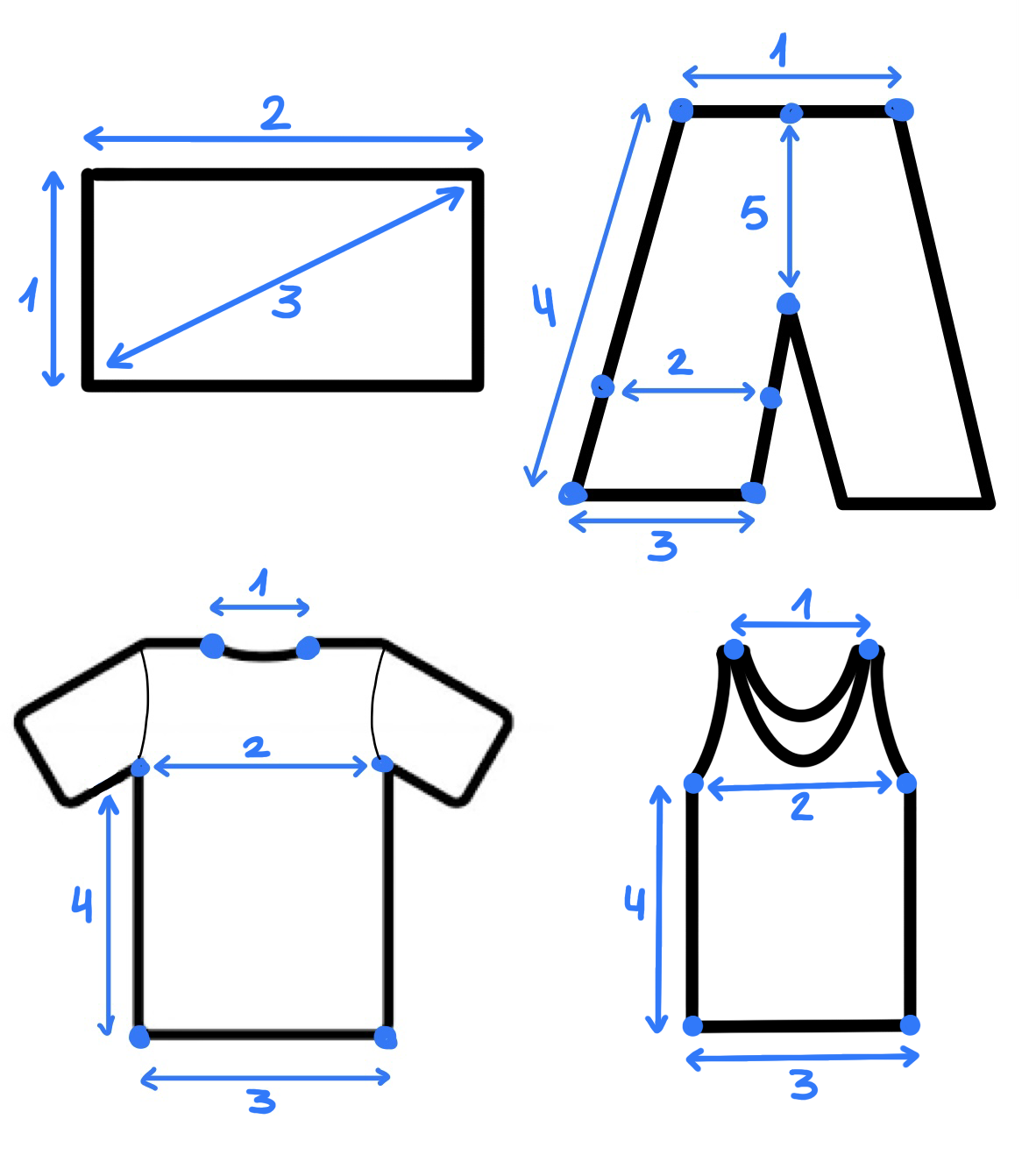}
        \caption{Reference lines for measuring size and elasticity.}
        \label{fig:elasticity}
\end{figure}

\textbf{Size:} The size of an object 
%constitutes one of the initial considerations in robotic manipulation, as it 
determines the necessary workspace to properly handle it.
%As cloth objects can have different shapes, we specify several lines where the length has to be measured. 
To measure different shapes, we define lines (see Fig. \ref{fig:elasticity}) according to their role in different tasks. For example, lines 1 and 3 of a T-shirt, top or pants are relevant for dressing, while line 4 is relevant for folding.

\textbf{Weight:} The mass of an object is generally considered taking into account the payload capacity of the robotic manipulator. The total mass of the object can be measured in Kilograms using a standard scale. 
\begin{wrap}
    \IG{specify precision of the scale?} 
\end{wrap}

\textbf{Shape:} The shape of the cloth provides information of its category and intended purpose. We propose to report the number of different shapes by classifying them into categories (e.g. rectangular clothes, shirts, skirts, pants, etc.).

\textbf{Color:} The colour of an object plays a significant role in perception algorithms. We propose to identify only the plain primary and secondary colours, to avoid entering into colour hue details and also differentiate prints. 

\textbf{Fabric material:} 
In previous works \cite{longhini2021textile} we showed how the material is relevant for robot manipulation, and therefore, should be taken into account. Examples include natural fibres such as cotton, linen, silk, and wool, as well as synthetic fibres like polyester, nylon, and acrylic.

\textbf{Construction technique:} The construction technique refers to the method or process used to create the fabric. The most common are woven and knitted, which can be identified by visual inspection. 
As shown in \cite{longhini2023elastic}, the construction technique significantly influences mechanical properties, such as elasticity.

\subsection{Mechanical properties}

The mechanical properties are parameters that describe how a material responds to applied forces or manipulations. They will depend on physical properties such as weight, fibre material or the fabric construction technique.

\textbf{Stiffness:} Cloth stiffness or rigidity influences how it behaves under manipulation as it determines the resistance to deformation. 
Understanding cloth stiffness can help in predicting cloth behaviour when manipulating and therefore improve planning tasks. 

Our goal is to design systems tailored for robotic manipulations, using whole objects without needing to cut them into standardized sizes. We seek to capture the behaviour of the cloth, including its inhomogeneities caused by added features such as buttons or hems. For example, in the case of two exact napkins of the same fabric and size, but where one of them has hems on the edges and the other one does not, we will see less stiffness with the hems due to heavier edges. This is expected and it is desirable to be detected, since during real manipulation executions, these two objects will behave differently under the same manipulations.

We propose a method inspired by the Cusick drape test \cite{cusick1968}, placing the object on a flat circular surface and measuring the draped cloth's area from above. However, instead of relying on specialized machines, we designed a perception algorithm that measures this area using zenithal images of the draped cloth and detecting its contour. 
To measure objects of different sizes without cutting cloth samples of standardize dimensions, we adapt the size of the flat surface used to place the cloth. In the original test, the 60\% of the cloth sample covered the plate, leaving the remaining 40\% of the fabric hanging and determining the stiffness. We maintain the same hanging cloth ratio by creating circular plates, whose diameter is determined by the shortest edge of the object. %For example, a towel measuring 10x40cm would require a 6cm diameter plate (60\% of 10cm).
Furthermore, to simplify the measurement process of objects with complex shapes (T-shirts or pants) or very big sizes (larger than 50cm), we use the objects folded in rectangular shapes. This standardization facilitates the area measurement as well as the whole assessment procedure.

\begin{figure}
\vspace{6pt}
    \centering
    \includegraphics[width=\linewidth]{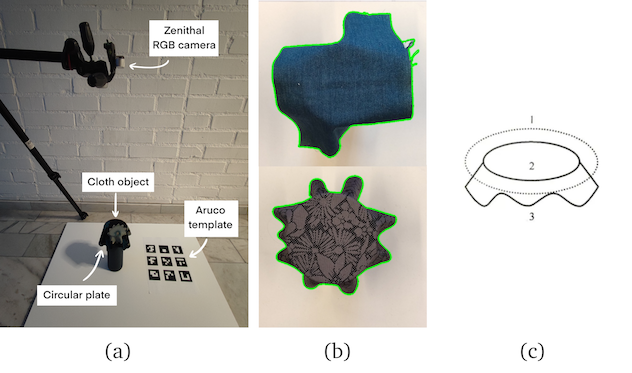}
    \caption{(a) Setup to measure stiffness, (b) examples of cloth drapeability and contour and (c) areas used for stiffness formula.}
    \label{fig:stiffness_areas}
    \vspace{-3pt}
\end{figure}

The formula used to measure the stiffness is the same as used in the original Cusick drape test:
\begin{equation}
    \text{\textit{stiffness}} = \frac{A_3-A_2}{A_1 - A_2}
\end{equation}
where $A_{1}$ is the initial area of the cloth, $A_{2}$ is the plate area used and $A_{3}$ is the area of the draped cloth given by the code (see Fig. \ref{fig:stiffness_areas}-c).

\textbf{Elasticity:} The elasticity or stretchability of clothes refers to the ability to extend its length when subjected to external forces. %and return to its original shape once this force is removed. 
It is an important property in dressing tasks, where it plays an important role in safety, as it helps in compensating movements of the user, reducing the risk of causing him harm. % and avoiding damaging the object. 
It is measured by pulling two sides of the cloth in opposite directions and assessing its elongation with: 
\begin{equation}
    \text{\textit{elasticity}} = \frac{l_f-l_i}{l_i}
\end{equation}
where $l_{i}$ is the length of the cloth between the subjected points at rest and $l_{f}$ is the length to which the cloth arrives while pulling.

We aim to standardise the measurement systems as much as possible with the goal of designing replicable and comparable systems. Thus, we propose the use of a luggage scale to apply a controlled tensile force of 0.5Kg. We also specify clamps as the way to grasp the fabric since the area of the contact point can influence in the measurement of elasticity.%\IG{image of setup (scale + clamp)?}

The stretchability of fabric depends on the construction technique and the direction of the yarn (warp, weft or bias). For this reason, we define several direction lines where to measure the elasticity, which are the same defined for measuring the size (Fig. \ref{fig:elasticity}).
%depending on whether it is being stretched in the warp or weft yarn directions.
%jersey knits are usually more elastic.
The type of fibre used in the textile plays a significant role. Synthetic fibres like spandex and elastane have inherently high elasticity, while natural fibres like cotton and linen have lower inherent elasticity. %Synthetic fibers like spandex and elastane are inherently more elastic than natural fibers like cotton or wool.
%The elasticity of a cloth is given by the material and construction technique. Generally, textiles with elastane have more elasticity.

\textbf{Friction:} The friction force of cloth can be defined as the resistance of the object to slide in contact with a surface. %It depends on several factors but mainly on the texture of the textile surface (roughness or the presence of fibers), and on the environmental surface used. For this reason, it influences in all those manipulations performed in contact with environmental objects, such as folding or unfolding of clothes in a table. 
%It is given mainly by the roughness or texture of the textile surface and depends on the type of environmental surface used. 
To measure the friction, we place the cloth object on a plane and gradually increase its angle of inclination by lifting one side of the plane until the sample starts to slide. In a similar way as with the stiffness system, for practical reasons, we fold objects of complex shapes or large sizes (such as T-shirts or tablecloths) to fit inside the surface. %In order to ensure reliable results, we recommend to repeat the experiment in reverse way, that is to say, placing the object on the inclined plane fixed at the previously obtained lifted height and observe if it slips. If it does, decrease the inclination lowering the height one centimetre and repeat the process. Repeat this until obtaining the height of the inclined plane edge (i.e. inclination angle) where the object does not slip. 
%For comparability purposes we use a paper. However, in order to have results in a specific environment it is usefull to do these measurement in the tackled surface.
We standardise the surface to have comparable values, specifying the use of a standard printing paper, which should have the necessary dimensions so that the object stays completely inside, and must be fixed to the selected surface where the test is going to be performed (e.g. a table).
We obtain the coefficient of friction ($\mu$) through trigonometry, knowing that the frictional force $\text{F}_{f}= \mu\text{N}= \mu\,\text{m}\text{g}\,cos(\theta)$, the force of gravity is ${\text{F}_{g}= \text{m}\text{g}sin(\theta)}$ and $\theta= sin^{-1}(\frac{h}{l})$, as:

\begin{equation}
    \text{\textit{friction}}=\mu=tan\left(sin^{-1}\left(\frac{h}{l}\right)\right)
\end{equation}

where $h$ is the reported height and $l$ is the length of the surface used.

We have selected this method to measure the friction since %its principle is directly related to the situations where the friction is going to influence the manipulation. This test 
it is valuable to assess the slip resistance of textiles, which is the main aspect that will influence cloth manipulations.  
In addition, this method has also been used previously in \cite{li2018model}. %To tune simulation parameters replicate the real behaviour of cloth.

Detailed guidelines of the systems with explanatory figures, resources and protocols can be found in the related website$^{\ref{fn:website}}$.

%%%%%%%%%%%%%%%

\begin{figure}[h]
\vspace{6pt}
        \centering
        \includegraphics[width=\linewidth]{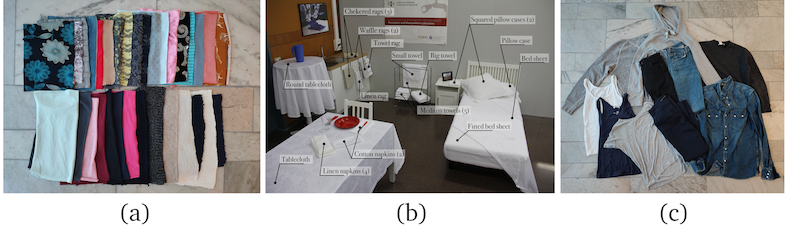}
        \caption{Cloth sets evaluated: (a) Elastic Object Set \cite{longhini2023edonet}, (b) Household Cloth Object Set\cite{irene2022household}, (c) Dressing Object Set \cite{gustavsson2022landmark}.}
        \label{fig:object_sets}
\end{figure}%

\subsection{Cloth set benchmarking} %Radar chart

In addition to characterizing objects by quantifying their properties, we can use the abovementioned measurement systems to benchmark various cloth sets, by identifying their strengths in terms of variability.
%to compare between different cloth sets to identify their strengths in terms of variability. This can be done by measuring and representing the variability of each property in a radar chart.
%Having the previous measures, we can build a radar chart to compare between different cloth sets by collecting the measures for different objects and obtaining the difference between the maximum and minimum values of each property. 
To do so, we can build a radar chart by collecting measures for various objects and computing the range between the maximum and minimum values of each property, determining the extent of variability. 
Thus, each axis will have different scales according to the units of the corresponding property.
This is a visual manner to compare which cloth sets have more diversity in each property, providing knowledge of in which applications an object set would be useful. For example, a cloth set with a lot of variation in colors or shapes is useful for evaluating the generalization of perception algorithms, while if it has a high range in stiffness or elasticity then is useful for manipulation purposes. 

As showcase, we gathered measures for all the objects of three different cloth sets found in the literature of cloth manipulation \cite{longhini2023edonet, irene2022household, gustavsson2022landmark} (see Fig. \ref{fig:object_sets}). We also use this representation to analyse the adequacy of them for the meant purposes as well as what strengths they have.
The first object set, which we have called Elastic Object Set (EOS), was created in \cite{longhini2023edonet} to train a model that learns the inherent elastic properties of fabrics through the exploration of pulling actions. %Specifically, it leverages a latent representation of elastic properties of textile objects through pulling actions. %understanding how different fabrics stretch, deform, and respond to forces.
It includes 37 rectangular textile samples of similar sizes. % \IG{with different degrees of elasticity}. %elastic behaviours.
The second object set, called Household Cloth Object Set (HCOS), was introduced in \cite{irene2022household} to foster benchmarking in cloth manipulation through the standarization of objects. It is composed of 27 household cloth objects with different sizes, shapes and features for performing different tasks. %, with the aim of allowing the development of a variety of household assistive tasks. 
The last object set, Dressing Object Set (DOS), was used in \cite{gustavsson2022landmark}. It comprises a set of 10 clothing items that were used for creating a dataset of 117 color images used to train a model for category classification and landmark detection.

From the resulting radar chart seen in Fig. \ref{fig:radar_chart}, which represents the variability of each measured property for the three object sets, we can see that EOS dataset has high variation in the three mechanical properties (stiffness, stretchability and friction), but low variability in size and weight. This is because the dataset was composed of small samples with low mass (around 10 gr on average) and similar size (up to 32cm). However, it was used for learning elastic properties, so we can say that this object set is effective for its purpose. 
It is also the dataset with more number of colors and materials, as it included samples from different clothing items of different prints and types of fibres. % objects with many different prints and made of different types of fibers, as it was created from different types of clothing objects.
By contrast, we see a dominance of the HCOS in size and weight, due to its variety of objects for household chores, including smaller items like towels for storing, as well as larger objects like bedsheets and tablecloths for bed-making or table-setting tasks. It also presents a high stiffness given that this mechanical property is highly dependent on these two physical properties.
%By contrast, we clearly see a predominance of the HCOS in size and weight. This is attributed to its diverse range of objects for different household chores, encompassing tasks that require smaller items like towels and napkins for folding and storing, as well as larger objects such as bedsheets and tablecloths for making beds or setting a table. % since it includes objects for different household chores such as folding towels, making beds or setting a table, which needs of small objects as towels or napkins and big items as bedsheets or tablecloths.
Regarding the DOS, it is composed of clothing items, which stand out for their varied shapes. %, so we can identify a peack in this property.
%it has a high variability in shape as it is composed of clothing items of different categories. for category classification. % different categories
It also has a high variation of stiffness and elasticity, although this was useless for its purpose (perception evaluation), due to the inclusion of denim objects, which are characterized for having a very low elasticity and high stiffness, and tank tops made of elastane and cotton, which usually provide low stiffness and high elasticity.
\begin{wrap}
    \IG{DOS has fewer materials but the same elasticity as EOS}
\end{wrap}

From this evaluation, we can say that creating a cloth set that covers a wide range in all properties is a complex task. For this reason, the proposed measurement systems can help in improving the creation of a cloth set that suits your purpose.

\vspace{5pt}

\section{Relevance of Cloth Properties in Robotic Manipulation }
%Evaluation via Action Primitives 
\label{sec:primitives} % (taxonomy) 

This section aims to explore the influence of the mechanical properties in the manipulation of CDO. Manipulation tasks can be divided into sequences of low-level actions~\cite{irene2022knowledge}, also referred to as primitives. We focus on the following set of quasi-static action primitives: lift, drag, fold, stretch and push.

 \subsection{Action Primitives}

\begin{figure}
\vspace{4pt}
        \centering
        \includegraphics[width=\linewidth]{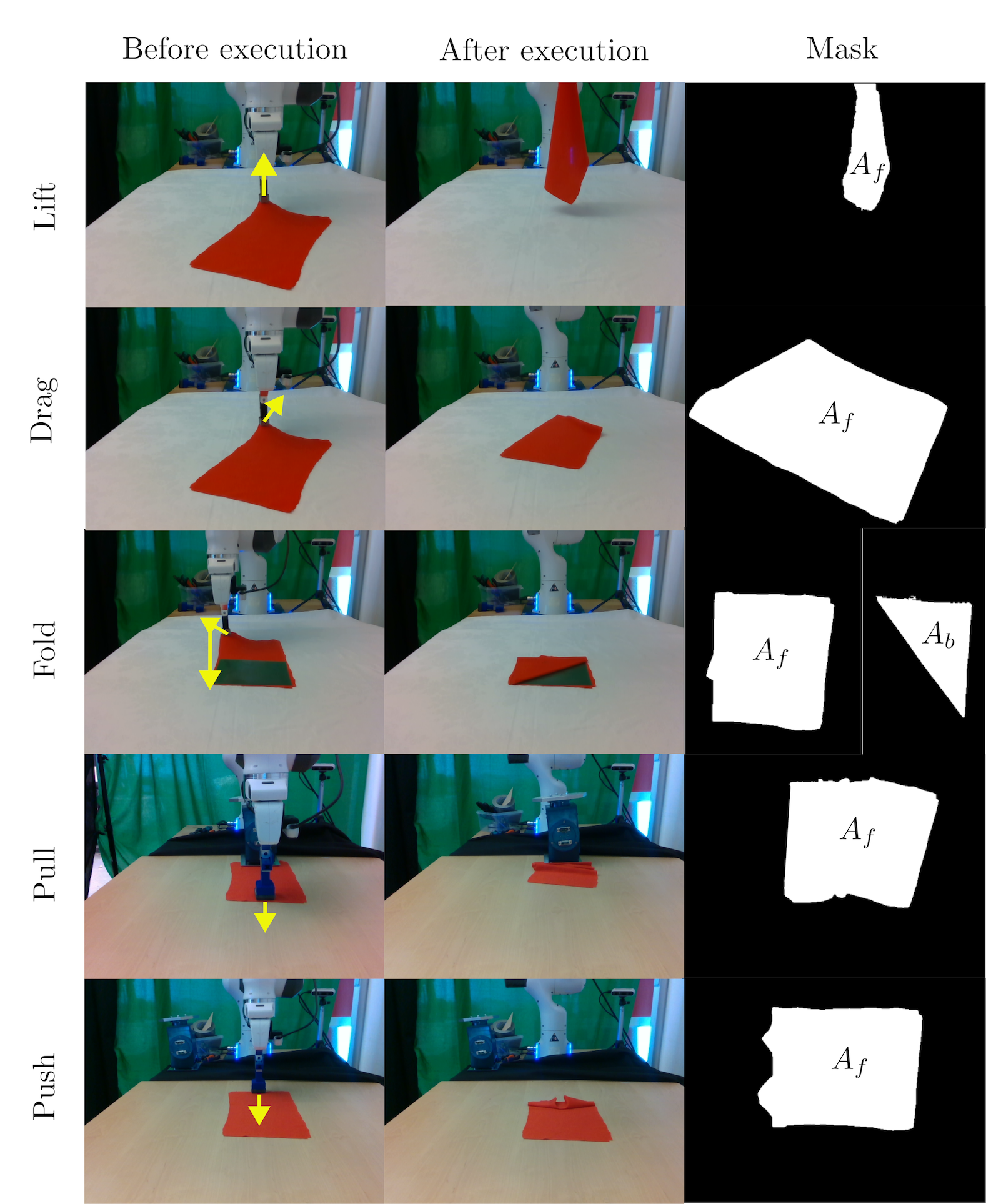}
        \caption{Example visualisations of the initial (left) and final states (middle) of each manipulation primitive (Lift, Drag, Fold, Pull, Push) and the mask of segmented cloth in the final state.}
        \label{fig:primitives}
        %\vspace{-3pt}
\end{figure}

We define each action primitive through the set of the initial pose, final pose, and the trajectory connecting the two poses, assuming that the cloth starts from a flattened configuration (see Fig. \ref{fig:primitives}). For the lift, drag, and fold primitives, we operate under the assumption that the cloth begins in a pre-grasped state. For the stretch and push primitives, we utilize a 3D printed end effector, which boasts a finger-like design akin to the one described in~\cite{qiu2023robotic}. We employ a Franka-Emika Panda robot for executing these primitives. Our primary objective is to consistently apply these action primitives across objects with varying properties, to discern how these properties impact manipulation outcomes. We define the action primitives as:

 \noindent \textbf{Lift:} From a flattened configuration where a corner of the cloth is grasped $3$cm above the table, the robot arm elevates the cloth until it is completely hanging, reaching a final position $35$cm higher than the starting point in a direction perpendicular to the table's surface, following a linear trajectory. 
 
  \noindent \textbf{Drag:} The top-left corner of the cloth is grasped $1$cm above the table, and the arm moves the cloth reaching a final position $20$cm further from the initial position in a direction parallel to the table's plane, following a linear trajectory. 
 
  \noindent \textbf{Fold:} The top-left corner of the cloth is grasped $3$cm above the table, and the arm reaches the final position, corresponding to the opposite corner of the cloth,  using a triangular path peaking at $11$cm at the midpoint between the start and end positions. 
 
  \noindent \textbf{Pull:}  We define the initial contact point with the cloth in the middle of the shorter edge of the cloth. We simultaneously fix the opposite side of the cloth with a heavy object to force the elastic behaviour. % to prevent it to slide during the execution. 
 The robot pulls the cloth to the final pose which is $5$cm further along the axis parallel to the table surface, following a linear trajectory.  
 
  \noindent \textbf{Push:}  We define the initial contact point with the cloth in the middle of the shorter edge of the cloth. The robot then pushes the cloth to the final pose which is $10$cm towards the centre of the cloth, following a linear trajectory. 

\begin{table*}
\vspace{6pt}
    \centering
    \caption{Results of the action primitive for 6 samples (A-B), characterized with their values obtained in the radar chart for the stiffness, elasticity and friction properties. The results of each primitive represent the Final Ratio (FR). In \textbf{bold} we represent the results with highest FR, corresponding to minimum deformation (or best fold for the Fold primitive), while we \underline{underline} the results showing the lowest FR, corresponding to the highest deformation (or worst fold).}
    \begin{tabular}{l|ccc|ccccccc}\toprule
          Sample & Stiffness & Elasticity                 &    Friction        &    Lift      &    Drag      &    Fold & Pull & Push                \\
        \toprule   
         A    & \textbf{85\%}      & 43\%             & 53\% & \boldsymbol{$ 0.31  \pm \; \; 0.01$}  & \boldsymbol{$0.97 \pm \; \; 0.01$}  & \boldsymbol{$1.00 \pm \; \; 0.0 $}   & $0.94  \pm \; \;  0.02 $   & $0.83  \pm \; \;  0.00$   \\ %\midrule
         B  &34\%               & \textbf{7\%}     & \textbf{45\%}  & $ 0.23 \pm \; \;   0.00 $  & $0.96  \pm \; \; 0.02$  & $0.63  \pm \; \;  0.01 $   &  \boldsymbol{$0.97  \pm \; \;  0.00$}   & \boldsymbol{$0.84  \pm \; \;   0.01$}   \\ %\midrule
         C    & 36\%               & 87\%             & 52\%    & $ 0.23  \pm \; \; 0.02 $  & $0.90 \pm \; \; 0.01$  & $0.63 \pm \; \;    0.00$ &\underline{ $  0.72  \pm \; \;  0.07 $}   &\underline{ $0.64  \pm \; \;    0.03 $ }  \\ %\midrule
         D   & 39\%               & 35\%             & \textbf{93\%}  & \underline{$ 0.20 \pm \; \;  0.00 $}  & $0.84  \pm \; \; 0.02$  & $ 0.63  \pm \; \;   0.01$ & $ 0.90  \pm \; \;    0.03$   & $0.69   \pm \; \;      0.05$   \\ %\midrule
        E    & 59\%               & \textbf{100\%}   & 60\%   & $ 0.21 \pm \; \;   0.01$  & $0.93 \pm \; \; 0.00$  & $  0.60  \pm \; \;   0.02$ & $  0.91 \pm \; \;     0.01$   & $0.65 \pm \; \;    0.02$   \\ %\midrule
        F     & \textbf{32\%}      & 64\%             & 52\%   & \underline{$0.20 \pm \; \;    0.00 $}  & \underline{$0.79  \pm \; \; 0.03$}  & \underline{$0.60   \pm \; \;  0.00$} & $ 0.88 \pm \; \; 0.02$   & $0.66 \pm \; \;    0.02$   \\ %\midrule
    \end{tabular}
    \label{tab:samples_results}
    %\vspace{-\baselineskip}
\end{table*}

 \subsection{Evaluation Metrics}\label{sec:evaluation_metrics}

To explore the influence of the cloth properties on the manipulation primitives, we focus on the mechanical properties (i.e. elasticity, stiffness, and friction) as the ones leading to different responses to applied forces or manipulations. Using the results of the measurement systems presented in the radar chart of Fig. \ref{fig:radar_chart}, we selected samples from the EOS dataset ensuring to include the extremes of each property. 

%%%%%%%%  %%%%%%%%%
We use shape-retention as the metric to evaluate the outcome of the drag, lift, push and pull primitives. This type of metric evaluates to what extent the initial shape of the cloth is maintained after the manipulation, starting from a flattened configuration. %\corrections{This type of metric is commonly used in literature to benchmark cloth manipulation employing various approaches, such as through contour extraction or image difference.} 
We measure shape-retention by computing the Final Ratio (FR) between the area covered by the cloth before ($\text{A}_{i}$) and after ($\text{A}_{f}$) the execution:

$$\text{FR}= \frac{\text{A}_{f}}{\text{A}_{i}}$$

Thus, a task will be considered successful when the cloth retains its shape after the execution, corresponding to a ${\text{FR}= 1}$. For the folding primitive, instead, we define as evaluation metric the alignment of the two halves of the cloth after the folding execution. We measure the alignment by evaluating the Final Ratio (FR) between the area covered by the top half of the cloth ($\text{A}_{t}$) and the final area covered by the entire cloth after execution ($\text{A}_{f}$), computed as ${\text{FR}= \text{A}_{t}/\text{A}_{f}}$. We obtain $\text{A}_{t}$ from the evaluation of the uncovered area of the bottom half of the cloth ($\text{A}_{b}$) as  $\text{A}_{t}= \text{A}_{f} - \text{A}_{b}$. Therefore, a value ${\text{FR}= 1}$ represents a good fold, corresponding to a perfect alignment of the two halves. We compute the areas for each metric by segmenting the cloth with~\cite{kirillov2023segment}, and counting the number of pixels of the segmentation mask (see right images of Fig. \ref{fig:primitives}).

\subsection{Results}

For the comparison among different mechanical properties, we selected 6 samples from the EOS dataset. We reported the outcome of each primitive in Table~\ref{tab:samples_results}, along with the characterisation of the samples measured with our framework. The results show that sample A, the one with the highest stiffness, is achieving the highest FR for the lift, drag and fold primitives. Sample B obtained the highest FR for the pull and push primitives being the one with the lowest friction, despite the low stiffness.  %However, sample A obtained similar results to sample B. These results suggest that stiffness influences shape retention across all primitives, while
Sample F has low stiffness and undergoes high deformations (lower FR) under the lift, drag and fold primitives. Sample D, on the other hand, due to the high elasticity and the low stiffness, deforms the most under the pull and push primitives. These results confirm that stiffness emerges as a crucial factor influencing shape retention across all primitives, while friction has a fundamental role  where there is contact between the cloth and the environment during the manipulation. Elasticity, instead, plays a pivotal role in scenarios where the cloth is under stress.

%\input{sections/Evaluation}

% =======================================
\section{Conclusions and Future work}

In this work, we introduced practical and scalable systems to measure cloth properties, designed for robotics applications. Our framework enables the characterization of cloth-like deformable objects by quantifying their physical and mechanical properties, and provides a structure to build and compare cloth sets. We compared three cloth sets from the literature, highlighting the challenge of creating a set with a broad range of properties. 
Furthermore, to improve understanding of garment manipulation approaches, works should include detailed descriptions of the objects used in the evaluations. These descriptions would clarify if the evaluation involves clothes with either narrow or broad sets of characteristics, indicating its relevance to the application. Our proposed framework, in particular, will be valuable for measuring these cloth descriptions and enhancing benchmarking in the cloth manipulation community.

We have further analysed the relevance of cloth mechanical properties showing how the outcome of the manipulations changes significantly for different properties.
This novel quantitative analysis highlights the impact of cloth properties on robotic manipulations, previously unexplored due to the absence of objective measurement systems. 

Our research directions will tend towards expanding this framework for performing local measurements of cloth parts. This enhancement will make possible to provide detailed information not only of the entire object but also of the parts that will be manipulated, capturing the inhomogeneities (e.g. buttons, zippers, etc) and enabling the robot to better understand the intricacies of fabric.
In addition, we plan to extend the presented study with more complex manipulations and objects, including different sizes and shapes, to identify more in-depth relations and effects and to explore the influence of the physical properties in robotic manipulations.

\normalem
\bibliographystyle{IEEEtran}
\bibliography{Final_version.bib}

\end{document}